\documentclass[journal=jcisd8,manuscript=article]{achemso}
\usepackage[T1]{fontenc} 
\usepackage{hyperref} 
\usepackage{amsmath} 
\usepackage{amsfonts}
\usepackage{algorithm}
\usepackage{algpseudocode} 
\usepackage{enumitem} 
\usepackage{booktabs}
\usepackage{siunitx}

\SectionNumbersOn

\author{Jaechang Lim}
\affiliation[KAIST]
{Department of Chemistry, KAIST, Daejeon, South Korea}
\altaffiliation{Contributed equally to this work}
\author{Sang-Yeon Hwang}
\affiliation[KAIST]
{Department of Chemistry, KAIST, Daejeon, South Korea}
\altaffiliation{Contributed equally to this work}
\author{Seungsu Kim}
\affiliation[KAIST]
{Department of Chemistry, KAIST, Daejeon, South Korea}
\author{Seokhyun Moon}
\affiliation[KAIST]
{Department of Chemistry, KAIST, Daejeon, South Korea}
\author{Woo Youn Kim}
\affiliation[KAIST]
{Department of Chemistry, KAIST, Daejeon, South Korea}
\alsoaffiliation[Second University]
{KI for Artificial Intelligence, KAIST, Daejeon, South Korea}
\email{wooyoun@kaist.ac.kr}

\title
  {Scaffold-based molecular design using graph generative model}

\begin{document}







\begin{abstract}
Searching new molecules in areas like drug discovery often starts from the core structures of candidate molecules to optimize the properties of interest. The way as such has called for a strategy of designing molecules retaining a particular scaffold as a substructure. On this account, our present work proposes a scaffold-based molecular generative model. The model generates molecular graphs by extending the graph of a scaffold through sequential additions of vertices and edges. In contrast to previous related models, our model guarantees the generated molecules to retain the given scaffold with certainty. Our evaluation of the model using unseen scaffolds showed the validity, uniqueness, and novelty of generated molecules as high as the case using seen scaffolds. This confirms that the model can generalize the learned chemical rules of adding atoms and bonds rather than simply memorizing the mapping from scaffolds to molecules during learning. Furthermore, despite the restraint of fixing core structures, our model could simultaneously control multiple molecular properties when generating new molecules.

\end{abstract}

\section{Introduction}
\label{sec:Introduction}
The ultimate goal of drug discovery is to find novel compounds with desirable pharmacological properties. Currently, it is an extremely challenging task, requiring years of development and numerous times of trials and failures.\cite{Hoelder2012} This is mainly because of the huge size and complexity of chemical space. For instance, the number of potential drug candidates is estimated about $10^{23}$ to $10^{60}$, whereas only $10^{8}$ molecules have ever been synthesized.\cite{Polishchuk2013, Kim2016} Furthermore, the discrete nature of molecules makes searching in chemical space even harder. 


Molecular generative models are attracting great attention as a promising in silico molecular design tool for assisting drug discovery.\cite{Chen2018, Sanchez-Lengeling2018} In previous works on molecular design, deep learning techniques with SMILES representation\cite{Weininger1988} of molecules were shown to be effective.\cite{gomez2018automatic, kang2019, lim2018, segler2017generating,Gupta2018,bjerrum2018improving,popova2018deep, Olivecrona2017,guimaraes2017objective,jaques2016,neil2018exploring,Polykovskiy2018}
Various deep generative models such as variational autoencoder\cite{gomez2018automatic, kang2019, lim2018}, language models\cite{segler2017generating,Gupta2018,bjerrum2018improving}, generative adversarial network,\cite{guimaraes2017objective} and adversarial autoencoder\cite{Polykovskiy2018} have been utilized to develop SMILES-based molecular generative models. 
The models have demonstrated their utility for generating new molecules and controlling their molecular properties. 

Despite the success of the SMILES-based molecular generative models, the SMILES has fundamental limitations in fully capturing molecular structures.\cite{Jin2018} Molecules having a high molecular similarity may have completely different SMILES representations. In this case, SMILES-based molecular generative models may fail to construct a smooth latent space. In addition, to generate valid SMILES strings, the models have to learn the grammar of SMILES, which increases the difficulty of the learning process and makes SMILES less preferable. 

In contrast to the SMILES, graphs can naturally express molecular similarity and validity. However, graph-based molecular generative models have been relatively less studied than SMILES-based models because generating meaningful graphs imposes more difficulties than generating sequences does. Recent improvement of graph generative models on such problem opens a new possibility of molecule generation with graph representations of molecules.

Li et al.\ proposed a graph generative model that predicts a sequence of graph building actions.\cite{Li2018} Starting from an empty graph, the model adds new nodes and edges in a successive manner. You et al.\ also developed a sequential graph generative model based on a deep auto-regressive model.\cite{you2018graph}
Another approach of generating graphs is to utilize generative adversarial nets.\cite{Iangoodfellow2014} Wang et al.\ developed GraphGAN, which is composed of a generator predicting node pairs that are most likely to be connected, and a discriminator classifying real connected pairs from fake ones.\cite{Wang2017}

Focusing on molecular design, Simonovsky et al. proposed GraphVAE, specialized for generating graphs of small molecules.\cite{simonovsky2018graphvae} GraphVAE directly generates an adjacency matrix, a node set, and an edge set of a fully-connected graph, and then recovers the original graph through a graph matching algorithm based on graph similarity. Jin et al. proposed a two-stage molecular graph generative model named JTVAE.\cite{Jin2018} In JTVAE, a graph tree structure composed of chemical substructures is constructed, and then the full graph of a molecule is produced by connecting the chemical substructures. Li et al.\ used a conditional graph generative model to design a dual inhibitor against c-Jun N-terminal kinase 3 and glycogen synthase kinase-3 beta, which are potential targets of Alzheimer's disease.\cite{li2018multi}

In real-world molecular design, a common strategy is first identifying initial candidates and then modifying their side chains while maintaining their scaffolds.
This strategy is particularly effective in designing protein inhibitors because the structural arrangement between a scaffold and protein residues is a main source of protein-ligand binding.
Despite its importance, such scaffold-based molecular design has drawn surprisingly less attention in developing molecular generative models.

One possible way of retaining a scaffold of generated molecules is searching the latent space constrained to the neighborhood of the scaffold.\cite{lim2018,Jin2018}
Alternatively, conditional generation of molecules can be used for controlling the scaffold of generated molecules by embedding the scaffold information in the latent space.\cite{li2018multi, Li2018}
However, those methods cannot assure with certainty that the generated molecules include a given scaffold as a substructure.
Also, the performance of the methods may vary depending on the types of scaffolds, or the methods may not be capable of incorporating scaffolds which were not in training set.
Furthermore, to our best knowledge, no work has shown to control the scaffold and the molecular properties of generated molecules at the same time.

In this regard, we developed a graph-based molecular generative model which can design new molecules having a given scaffold as a substructure. Our primary contribution is proposing a scheme of generating molecular graphs  from the graph of a given scaffold.
Our model extends a scaffold graph by sequentially adding new nodes and edges.
As a result, the model guarantees the generated molecules to include a given scaffold as a substructure.
Our model can generate molecules from arbitrary scaffolds, whether they were used for training the model or not.
This shows that our model can actually learn to build new molecules rather than memorize the patterns between the molecules and their scaffolds in the training dataset.

We tested whether our model can generate molecules with desirable properties under the constraint of fixing a scaffold.
Conditional molecule generation has been already reported in other molecular generative models.\cite{kang2019,lim2018,Jin2018,li2018multi}
However, property control becomes more challenging in scaffold-based molecule generation because fixing a scaffold confines chemical space, decreasing the possibility of finding desirable molecules.
Nevertheless, our model can efficiently control molecular properties with high validity and success rate of molecule generation.
Moreover, the model can control multiple properties simultaneously with comparable performance to the single-property result.

\section{Method}
\label{sec:Method}

\begin{table}
    \centering
    \begin{tabular}{l|l}
        \hline
        \textbf{Notation} & \textbf{Description} \\ \hline
        $G$ & An arbitrary or whole-molecule graph, depending on the context\\ \hline
        $S$ & A molecular scaffold graph \\ \hline
        $V(G)$ & The node set of a graph $G$ \\ \hline
        $E(G)$ & The edge set of a graph $G$ \\ \hline
        $\mathbf{h}_v$ & A node feature vector \\ \hline
        $\mathbf{h}_{uv}$ & An edge feature vector \\ \hline
        $\mathbf{H}_{V(G)}$ & $\left\{ \mathbf{h}_v: v\in V(G) \right\}$ \\ \hline
        $\mathbf{H}_{E(G)}$ & $\left\{ \mathbf{h}_{uv}: (u,v)\in E(G) \right\}$ \\ \hline
        $\mathbf{h}_G$ & A readout vector summarizing $\mathbf{H}_{V(G)}$ \\ \hline
        $\mathbf{z}$ & A lantent vector to be decoded \\ \hline
        $\mathbf{y}$ & The vector of molecular properties of a whole-molecule \\ \hline
        $\mathbf{y}_S$ & The vector of molecular properties of a scaffold \\ \hline
    \end{tabular}
    \caption{Notations used throughout the paper.}
    \label{tbl:notations}
\end{table}

\textbf{Overall process and model architecture.}
Our purpose is to generate molecules with target properties while retaining a given scaffold as a substructure. To this end, we set our generative model to be such that accepts a graph representation $S$ of a molecular scaffold and generates a graph $G$ that is a \textit{supergraph} of $S$. The underlying distribution of $G$ can be expressed as $p(G;S)$. Our notation here intends to manifest the particular relation, i.e., the supergraph--subgraph relation, between $G$ and $S$. We also emphasize that $p(G;S)$ is a distribution of $G$ alone; $S$ acts as a parametric argument, explicitly confining the domain of the distribution.
Molecular properties are introduced as a condition, by which the model can define conditional distributions $p(G;S|\mathbf{y},\mathbf{y}_S)$, where $\mathbf{y}$ and $\mathbf{y}_S$ are the vectors containing the property values of a molecule and its scaffold, respectively. Often in other works of molecule generation \cite{li2018multi, Li2018}, a substructure moiety is imposed as a condition, hence defining a conditional distribution $p(G|S)$. In such case, the distribution can have nonzero probabilities on graphs that are not supergraphs of $S$. On the other hand, the molecules that our model generates according to $p(G;S)$ always include $S$ as a substructure.
Before we proceed further, we refer the reader to Table~\ref{tbl:notations} for the notations we will use in what follows. Also, when clear distinction is necessary, we will call a molecule a ``whole-molecule'' to distinguish it from a scaffold.

The learning object of our model is a strategy of extending a graph to larger graphs whose distribution follows that of real molecules. We achieve this by training our model to recover the molecules in a dataset from their scaffolds. The scaffold of a molecule can be defined in a deterministic way such as that by Bemis and Murcko\cite{Bemis1996}, which is what we used in our experiments. The construction of a target graph is done by making successive decisions of node and edge additions. The decision at each construction step is drawn from the node features and edge features of the graph at the step. The node features and edge features are recurrently updated to reflect the construction history of the previous steps. The construction process will be further detailed below.

\begin{figure}[htb!]
    \centering
    \includegraphics[width=1.0\textwidth]{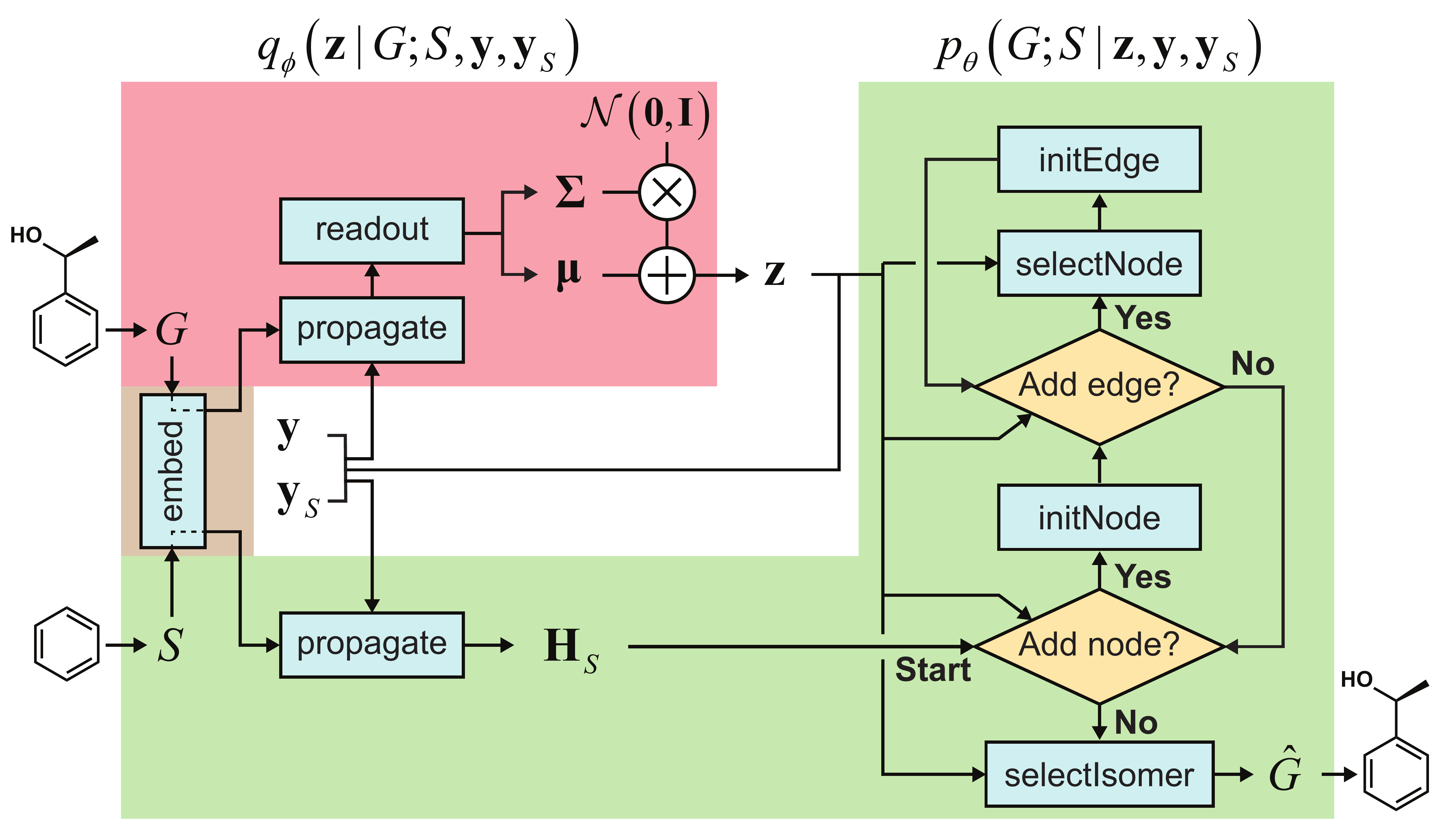}
    \caption{The model architecture in the learning phase. The encoder $q_\phi$ is trained to encode a whole-molecule graph $G$ into a latent vector $\mathbf{z}$, and the decoder $p_\theta$ is trained to recover $G$ from $\mathbf{z}$ by sequentially adding nodes and edges to the scaffold graph $S$. The modules in the red area constitute the encoder, and those in the green area constitute the decoder. In the generation phase after learning, only a scaffold is given, and $\mathbf{z}$ is sampled from the standard normal distribution $\mathcal{N}\left(\mathbf{0},\mathbf{I}\right)$. The whole process can be conditioned by molecular properties (expressed as $\mathbf{y}$ and $\mathbf{y}_S$).}
    \label{fig:architecture}
\end{figure}

We realized our model as a variational autoencoder (VAE),\cite{kingma2013auto} with the architecture depicted in Figure~\ref{fig:architecture}.
The architecture consists of an encoder $q_\phi$ and a decoder $p_\theta$, parametrized by $\phi$ and $\theta$, respectively. The encoder encodes a graph $G$ to an encoding vector $\mathbf{z}$, and the decoder decodes $\mathbf{z}$ to recover $G$.
The decoder requires a scaffold graph $S$ as an additional input, and the actual decoding process runs by sequentially adding nodes and edges to $S$. The encoding vector $\mathbf{z}$ plays its role by consistently affecting its information in updating the node and edge features of a transient graph being processed. Similarly to $p\left(G;S\right)$, our notation $p_\theta\left(G;S\vert\mathbf{z}\right)$ indicates that candidate generations of the decoder are always a supergraph of $S$. As for the encoder notation $q_\phi\left(\mathbf{z}\vert G;S\right)$, we emphasize that the encoder also has a dependence on the scaffold because of the joint optimization of $q_\phi$ and $p_\theta$.

\textbf{Graph encoding.}
The goal of graph encoding is to generate a latent vector $\mathbf{z}$ of the entire graph $G$ of a whole-molecule. Given the graph $G= \left(V(G), E(G)\right)$ of any whole-molecule, we first associate each node $v\in V(G)$ with a node feature vector $\mathbf{h}_v$ and each edge $(u,v)\in E(G)$ with an edge feature vector $\mathbf{h}_{uv}$. For the initial node and edge features, we choose the atom types and bond types of the molecule. We then embed the initial feature vectors in new vectors with a higher dimension so that the vectors have enough capacity to express deep information in and between the nodes and edges. To fully encode the structural information of the molecule, we want every node embedding vector $\mathbf{h}_v$ to contain not only the sole information of its own node $v$ but also the relation of $v$ to its neighborhood. This can be done by propagating each node's information to the other nodes in the graph. A large variety of related methods have been devised, each being a particular realization of a graph neural network.

In this work, we implemented the encoder $q_\phi$ as a variant of the interaction network\cite{Battaglia2016,gilmer2017}. Our network's algorithm consists of a propagation phase and a readout phase, which we write as
\begin{align}
    \label{eq:mpnn propagate}
    \mathbf{H}_{V(G)}' &= \mathsf{propagate} \left(\mathbf{H}_{V(G)}, \mathbf{H}_{E(G)}\right) \\
    \label{eq:mpnn readout}
    \mathbf{h}_{G} &= \mathsf{readout} \left(\mathbf{H}_{V(G)}' \right).
\end{align}
The propagation phase itself consists of two stages. The first stage calculates an aggregated message between each node and its neighbors as
\begin{equation}
    \label{eq:mpnn message}
    \mathbf{m}_v = \sum_{u:(u,v)\in E(G)} M\left(\mathbf{h}_u, \mathbf{h}_v, \mathbf{h}_{uv}\right) \quad\forall v\in V(G)
\end{equation}
with a message function $M$. The second stage updates the node vectors using the aggregated messages as
\begin{equation}
    \label{eq:mpnn update}
    \mathbf{h}'_v = U\left(\mathbf{m}_v, \mathbf{h}_v\right) \quad\forall v\in V(G)
\end{equation}
with an update function $U$. Updating every node feature vector in $\mathbf{H}_{V(G)}$ results in an updated set $\mathbf{H}'_{V(G)}$, as written in Eq.~\ref{eq:mpnn propagate}.
We iterate the propagation phase a fixed number of times whenever applied, using different sets of parameters at different iteration steps.
After the propagation, the readout phase (Eq.~\ref{eq:mpnn readout}) computes a weighted sum of the node feature vectors, generating one vector representation $\mathbf{h}_G$ that summarizes the graph as a whole. Then finally, a latent vector $\mathbf{z}$ is sampled from a normal distribution whose mean and variance are inferred from $\mathbf{h}_G$.

The graph propagation can be conditioned by incorporating an additional vector $\mathbf{c}$ in calculating aggregated messages. In such case, the functions $M$ and (accordingly) $\mathsf{propagate}$ accept $\mathbf{c}$ as an additional argument (i.e., they become $M\left(\cdot,\cdot,\cdot,\mathbf{c}\right)$ and $\mathsf{propagate}\left(\cdot,\cdot,\mathbf{c}\right)$). When encoding input graphs, we choose $\mathbf{c}$ to be the concatenation of the property vectors $\mathbf{y}$ and $\mathbf{y}_S$ to enable property-controlled generation. During graph \textit{decoding}, we use the concatenation of $\mathbf{y}$, $\mathbf{y}_S$, and the latent vector $\mathbf{z}$ as the condition vector (see below).

\textbf{Graph decoding.}
The goal of graph decoding is to reconstruct the graph $G$ of a whole-molecule from the latent vector $\mathbf{z}$ sampled in the graph encoding phase. Our graph decoding process is motivated by the sequential generation strategy of Li \latin{et al.}\cite{Li2018}
In our work, we build the whole-molecule graph $G$ from the scaffold graph $G_0$ (extracted from $G$ by the Bemis-Murcko method\cite{Bemis1996}) by successively adding nodes and edges. Here, $G_0=S$ denotes the initial scaffold graph, and we will write $G_t$ to denote any transient (or completed) graph constructed from $G_0$.

Our graph decoding starts with preparing and propagating the initial node features of $G_0$. As we do for $G$, we prepare the initial feature vectors of $G_0$ by embedding the atom types and bond types of the scaffold molecule. This initial embedding is done by the same network ($\mathsf{embed}$ in Figure~\ref{fig:architecture}) used for whole-molecules. The initial feature vectors of $G_0$ are then propagated a fixed number of times by another interaction network.
As the propagation finishes, the decoder extends $G_0$ by processing it through a loop of node additions and accompanying (inner) loops of edge additions. A concrete description of the process is as follows:
\begin{enumerate}[label=Stage \arabic*:]
    \item node addition. Choose an atom type or terminate the building process with estimated probabilities. If an atom type is chosen, add a new node, say $w$, with the chosen type to the current transient graph $G_t$ and proceed to Stage 2. Otherwise, terminate the building process and return the graph.
    \item edge addition. Given the new node, choose a bond type or return to Stage 1 with estimated probabilities. If a bond type is chosen, proceed to Stage 3.
    \item node selection. Select a node, say $v$, from the existing nodes except $w$ with estimated probabilities. Then, add a new edge $(v,w)$ to $G_t$ with the bond type chosen in Stage 2. Continue the edge addition from Stage 2.
\end{enumerate}
The flow of the whole process is depicted in the right side of Figure~\ref{fig:architecture}. Excluded from Stages 1--3 is the final stage of selecting a proper isomer, about which we describe separately below.

In every stage, the model draws an action by estimating a probability vector on candidate actions. Depending on whether the current stage should add an atom or not (Stage 1), add an edge or not (Stage 2), or select an atom to connect (Stage 3), the probability vector is computed by the corresponding one among the following:
\begin{align}
    \label{eq:addNode}
    &\hat{\mathbf{p}}^{an} = \mathsf{addNode} \left( \mathbf{H}_{V(G_t)}, \mathbf{H}_{E(G_t)}, \mathbf{z} \right) \\
    \label{eq:addEdge}
    &\hat{\mathbf{p}}^{ae} = \mathsf{addEdge} \left( \mathbf{H}_{V(G_t)}, \mathbf{H}_{E(G_t)}, \mathbf{z} \right) \\
    \label{eq:selectNode}
    &\hat{\mathbf{p}}^{sn} = \mathsf{selectNode} \left( \mathbf{H}_{V(G_t)}, \mathbf{H}_{E(G_t)}, \mathbf{z} \right).
\end{align}
The first probability vector $\hat{\mathbf{p}}^{an}$ is a $(n_a+1)$-length vector, where its elements $\hat{p}^{an}_1$ to $\hat{p}^{an}_{n_a}$ correspond to the probabilities on $n_a$ atom types, and $\hat{p}^{an}_{n_a+1}$ is the termination probability.
As for $\hat{\mathbf{p}}^{ae}$, a vector of size $n_b+1$, its elements $\hat{p}^{ae}_1$ to $\hat{p}^{ae}_{n_b}$ correspond to the probabilities on $n_b$ bond types, and $\hat{p}^{ae}_{n_b+1}$ is the probability of stopping edge addition.
Lastly, the $i$-th element of the third vector $\hat{\mathbf{p}}^{sn}$ is the probability of connecting the $i$-th existing node with the lastly added one.

When the model decides to add a new node, say $w$, a corresponding feature vector $\mathbf{h}_w$ should be added to $\mathbf{H}_{V(G_t)}$. To that end, the model prepares an initial feature vector $\mathbf{h}^0_w$ by representing the atom type of $w$ and then incorporates it with the existing node features in $\mathbf{H}_{V(G_t)}$ to compute a proper $\mathbf{h}_w$. Similarly, when a new edge, say $(v,w)$, is added, the model computes $\mathbf{h}_{vw}$ from $\mathbf{h}^0_{vw}$ and $\mathbf{H}_{V(G_t)} \cup \mathbf{h}_w$ to update to $\mathbf{H}_{E(G_t)}$, where $\mathbf{h}^0_{vw}$ represents the bond type of $(v,w)$. The corresponding modules for initializing new nodes and edges are as follows:
\begin{align}
    \label{eq:initNode}
    \mathbf{h}_w &= \mathsf{initNode} \left( \mathbf{h}^0_w, \mathbf{H}_{V(G_t)} \right) \\
    \label{eq:initEdge}
    \mathbf{h}_{vw} &= \mathsf{initEdge} \left( \mathbf{h}^0_{vw}, \mathbf{H}_{V(G_t)} \cup \mathbf{h}_w \right).
\end{align}

The graph building modules $\mathsf{addNode}$, $\mathsf{addEdge}$, and $\mathsf{selectNode}$ include a preceding step of propagating node features. For instance, the actual operation done in $\mathsf{addNode}$ is
\begin{equation}
    \label{eq:addNode2}
    \mathsf{addNode} \left( \mathbf{H}_{V(G_t)}, \mathbf{H}_{E(G_t)}, \mathbf{z} \right) = f\circ \mathrm{concat}\Big( \mathsf{readout} \circ \mathsf{propagate}^{(k)} \left( \mathbf{H}_{V(G_t)}, \mathbf{H}_{E(G_t)}, \mathbf{z} \right), \mathbf{z} \Big),
\end{equation}
where $\circ$ denotes the function composition. According to the right-hand side, the module updates node feature vectors through $k$ times of graph propagation, then computes a readout vector, then concatenates it with $\mathbf{z}$, and finally outputs $\hat{\mathbf{p}}^{an}$ through a multilayer perceptron $f$. Likewise, both $\mathsf{addEdge}$ and $\mathsf{selectNode}$ start with iterated applications of $\mathsf{propagate}$. In this way, node features are recurrently updated every time the transient graph evolves, and the prediction of every building event becomes dependent on the history of the preceding events.

As shown in Eq.~\ref{eq:addNode2}, the graph propagation in $\mathsf{addNode}$ (and $\mathsf{addEdge}$ and $\mathsf{selectNode}$) incorporates the latent vector $\mathbf{z}$, which encodes a whole-molecule graph $G$. This makes our model refer to $\mathbf{z}$ while making graph building decisions and ultimately reconstruct $G$ \textit{by decoding} $\mathbf{z}$. If the model is to be conditioned on whole-molecule properties $\mathbf{y}$ and scaffold properties $\mathbf{y}_S$, one can understand Eqs.~\ref{eq:addNode}--\ref{eq:selectNode} and \ref{eq:addNode2} as incorporating $\tilde{\mathbf{z}} = \mathrm{concat}\left( \mathbf{z}, \mathbf{y}, \mathbf{y}_s \right)$ instead of $\mathbf{z}$.

\textbf{Molecule generation.} When generating new molecules, one needs a scaffold $S$ as an input, and a latent vector $\mathbf{z}$ is sampled from the standard normal distribution. Then the decoder generates a new molecular graph $\hat{G}$ as a supergraph of $S$. If one desires to generate molecules with designated molecular properties, the corresponding property vectors $\mathbf{y}$ and $\mathbf{y}_S$ should be provided to condition the building process.

\textbf{Isomer selection.} Molecules can have stereoisomers, which have the same connectivity between atoms but different three-dimensional geometries. Consequently, the complete generation of a molecule should also specify the molecule's stereoisomerism. We determine the stereochemical configuration of atoms and bonds after a molecular graph $\hat{G}$ is constructed from $\mathbf{z}$.\cite{Jin2018} The isomer selection module $\mathsf{selectIsomer}$ prepares the graphs $I$ of all possible stereoisomers, enumerated by the RDKit software,\cite{landrum2006rdkit} whose two-dimensional structures without stereochemical labels are the same as that of $\hat{G}$. All the prepared $I$ include the stereochemical configuration of atoms and bonds in the node and edge features. Then the module estimates the selection probabilities as
\begin{equation}
    \label{eq:selectIsomer}
    \hat{\mathbf{p}}^{si} = \mathsf{selectIsomer}\left(\hat{G}, \mathbf{z}\right),
\end{equation}
where the elements of the vector $\hat{\mathbf{p}}^{si}$ are the estimated probabilities of selecting respective $I$.

\textbf{Objective function.}
Our objective function has a form of the log-likelihood of an ordinary VAE:
\begin{equation}
    \label{eq:objective}
    \log p\left(G;S\right) \ge \mathbb{E}_{\mathbf{z}\sim q_\phi} \left[\log p_\theta \left(G;S\vert\mathbf{z}\right)\right] - D_\mathrm{KL} \left[q_\phi \left(\mathbf{z}\vert G;S\right) \Vert p\left(\mathbf{z}\right)\right],
\end{equation}
where $D_\mathrm{KL}\left[\cdot\Vert\cdot\right]$ is the Kullback-Leibler divergence, and $p(\mathbf{z})$ is the standard normal prior.
In actual learning, we have a scaffold dataset $\mathcal{S}$, and for each scaffold $S\in\mathcal{S}$ we have a corresponding whole-molecule dataset $\mathcal{D}(S)$.
Note that any set of molecules can produce a scaffold set and a collection of whole-molecule sets: once the scaffolds of all molecules in a pregiven set are defined, producing $\mathcal{S}$, the molecules of the set can be grouped into the collection $\mathcal{D}(\mathcal{S}) = \left\{ \mathcal{D}(S): S\in\mathcal{S} \right\}$.
Using those datasets, our objective is to find the optimal values of the parameters $\phi$ and $\theta$ that maximize the right-hand side of Eq.~\ref{eq:objective}, hence maximizing $\mathbb{E}_{S\sim \mathcal{S}} \mathbb{E}_{G\sim \mathcal{D}(S)} \left[\log p\left(G;S\right)\right]$.

In \hyperref[sec:appendix]{Appendix} we detail our implementation of the modules and their exact operations. We also detail the full process of the model in Algorithm~\ref{alg:ggm}.

\section{Results and discussion}
\label{sec:results and discussion}

\subsection{Datasets and experiments}
\label{sec:datasets and experiments}
We obtained our dataset from the molecular library (version March 2018) provided by InterBioScreen Ltd. The raw dataset contained the SMILES strings of organic compounds composed of H, C, N, O, F, P, S, Cl, and Br atoms. We filtered out the strings containing disconnected ions or fragments and those that cannot be read by RDKit. Our preprocess resulted in \num{349726} training molecules and \num{116576} test molecules. The number of heavy atoms was 27 on average with a maximum of 132, and the average molecular weight was \SI{389}{\gram\per\mole}. The number of scaffold kinds was \num{85318} in the training set and \num{42751} in the test set.

Our experiments include the training and evaluation of our scaffold-based graph generative model using the stated dataset. For the conditional molecule generation, we used molecular weight (MW), topological polar surface area (TPSA), and octanol--water partition coefficient (LogP). We used one, two, or all of the three properties to singly or jointly condition the model. We set the learning rate to \num{0.0001} and trained all instances of the model up to 20 epochs. The other hyperparameters such as the layer dimensions are stated in \hyperref[sec:appendix]{Appendix}. We used RDKit to calculate the properties of molecules. In what follows, we will omit the units of MW (\si{\gram\per\mole}) and TPSA (\si{\square\angstrom}) for simplicity.

\subsection{Validity, uniqueness, and novelty analysis}
\label{sec:overall performance}

The validity, uniqueness, and novelty of generated molecules are basic evaluation metrics of molecular generative models. For the exact meanings of the three metrics, we conform to the following definitions:
\begin{gather*}
    \text{validity} = \frac{\text{\# of valid graphs}}{\text{\# of generated graphs}} \\
    \text{uniqueness} = \frac{\text{\# of nonduplicative, valid graphs}}{\text{\# of valid graphs}} \\
    \text{novelty} = \frac{\text{\# of unique graphs not in the training set}}{\text{\# of unique graphs}},
\end{gather*}
where we define a graph to be valid if it satisfies basic chemical requirements such as valency. In practice, we use RDKit to determine the validity of generated graphs.
It is particularly important for our model to check the metrics above because generating molecules from a scaffold restricts the space of candidate products.
We evaluated the models that are singly conditioned on MW, TPSA, or LogP by randomly selecting 100 scaffolds from the dataset and generating 100 molecules from each.
The target values (100 for each property) were randomly sampled from each property's distribution over the dataset.
For MW, generating molecules whose MW is smaller than the MW of its scaffold is unnatural, so we excluded those cases from our evaluation.

Table~\ref{tbl:validty_uniqueness_novelty} summarizes the validity, uniqueness, and novelty of the molecules generated by our models and the results of other molecular generative models for comparison.
Note that the comparison here is only approximate because the models were trained by different datasets.
Despite the strict restriction imposed by scaffolds, our models show high validity, uniqueness, and novelty, comparable to those of the other molecular generative models.
The high uniqueness and novelty are particularly meaningful considering the fact that most of the scaffolds in our training set have only a few whole-molecules. 
For instance, among the \num{85318} scaffolds in the training set, \num{79700} scaffolds have less than ten whole-molecules.
Therefore, it is unlikely that our model achieved such a high performance by simply memorizing the training set, and we can conclude that our model learns the chemical rules general in extending arbitrary scaffolds.

\begin{table}[htb!]
  \caption{Validity, uniqueness, and novelty of the molecules generated by our model, and the results from other molecular generative models.}
  \label{tbl:validty_uniqueness_novelty}
  \begin{tabular}{@{}l S[table-format=3.1] S[table-format=2.1] S[table-format=2.1]@{}}
    \toprule
    Model & {Validity} & {Uniqueness} & {Novelty} \\
    & {(\%)} & {(\%)} & {(\%)} \\
    \midrule
    Ours (MW) & 98.3 & 83.2 & 98.7   \\
    Ours (TPSA) & 93.7 & 84.4 & 99.1  \\
    Ours (LogP) & 97.1 & 88.0 & 99.2  \\ \addlinespace
    GraphVAE\cite{DeCao2018}  & 55.7 & 87.0 & 61.6  \\
    MolGAN\cite{DeCao2018}  & 98.1 & 10.4 & 94.2  \\
    JTVAE\cite{Jin2018}  & 100.0 & {--} & {--}  \\
    MolMP\cite{li2018multi}  & {95.2--97.0} & {--} & {91.2--95.1}  \\
    SMILES VAE\cite{li2018multi}  & 80.4 & {--} & 79.3  \\
    SMILES RNN\cite{li2018multi}  & 93.2 & {--} & 89.9  \\
    \bottomrule
  \end{tabular}
\end{table}

\subsection{Single-property control}
\label{sec:single-property control}

For the next analysis, we tested whether our scaffold-based graph generative model can generate molecules having a specific scaffold and desirable properties simultaneously.
Although several molecular generative models have been developed for controlling molecular properties of generated molecules, it would be more challenging to control molecular properties under the constraint imposed by a given scaffold.
We set the target values as 80, 100, and 120 for MW, 300, 350, and 400 for TPSA, and 5, 6, and 7 for LogP.
For all the nine cases, we used the same 100 scaffolds used for the result in Sec.~\ref{sec:overall performance} and generated 100 molecules for each scaffold.

\begin{figure}[htb!]
    \centering
    \includegraphics[width=0.5\textwidth]{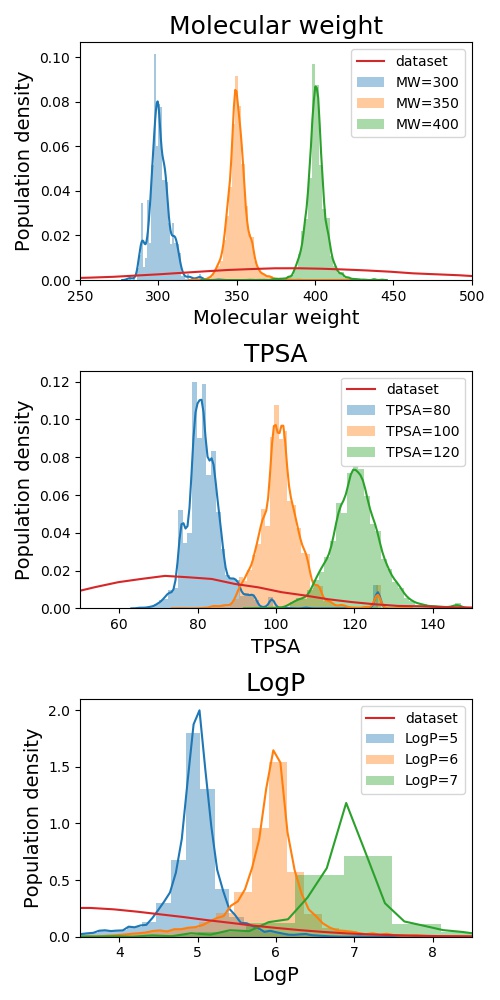}
    \caption{Property distributions of the generated molecules. The values in the legends indicate the target property values of the generation tasks. The red line in each plot shows the respective property distribution of the molecules in the training dataset.}
    \label{fig:single_prop_control}
\end{figure}

\begin{figure}[htb!]
    \centering
    \includegraphics[width=1.0\textwidth]{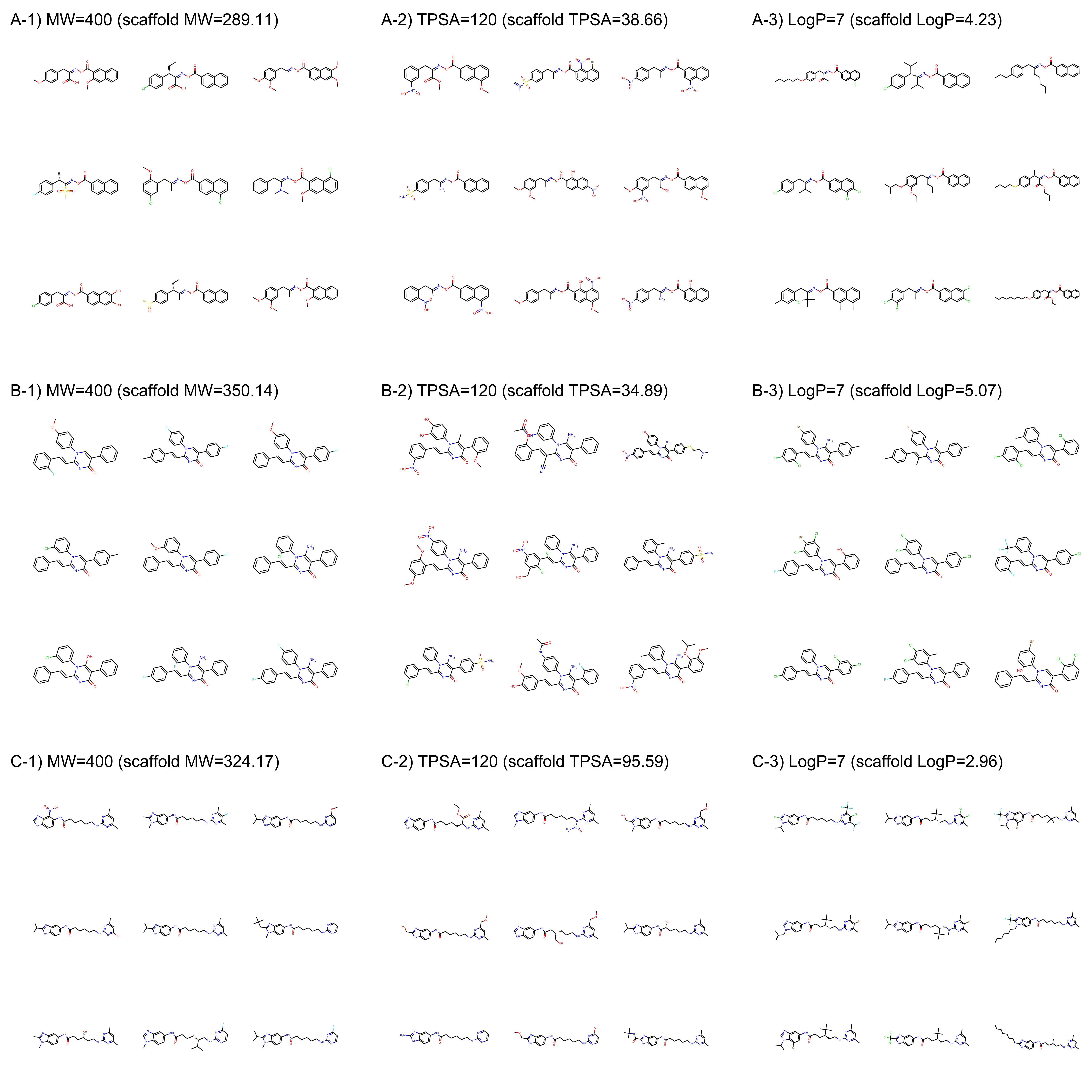}
    \caption{Example molecules generated from three scaffolds. The indicated values are the target conditions of the generation and the property values of the scaffolds.}
    \label{fig:example_molecules}
\end{figure}

Figure~\ref{fig:single_prop_control} shows the property distributions of generated molecules.
We see that the property distributions are well centered around the target values.
This shows that despite the narrowed search space, our model successfully generated new molecules with desirable properties.
To see how our model extends a given scaffold according to designated property values, we drew some of the generated molecules in Figure~\ref{fig:example_molecules}.
For the target conditions MW = 400, TPSA = 120, and LogP = 7, we sampled nine random examples using three different scaffolds.
The molecules in each row were generated from the same scaffold.
We see that the model generates new molecules with designated properties by adding proper side chains: for instance, the model added hydrophobic groups to the scaffolds to generate high-LogP molecules, while it added polar functional groups to generate high-TPSA molecules.

\subsection{Scaffold dependence}
\label{sec:scaffold dependence}
Our molecular design process starts from a given scaffold with sequentially adding nodes and edges. So the performance of our model can be affected by the kind of scaffolds.
Accordingly, we tested whether our model retains its performance of generating desirable molecules when new scaffolds are given.
Specifically, we prepared a set of 100 new scaffolds (henceforth ``unseen'' scaffolds) that were not included in the training set and an additional set of 100 scaffolds (henceforth ``seen'' scaffolds) from the training set.
We then generated 100 molecules for each scaffold with randomly designated property values. The process is repeated for MW, TPSA, and LogP.

Table~\ref{tbl:scaffold_dependence} summarizes the validity, uniqueness, and MAD of molecules generated from the seen and unseen scaffolds. Here, MAD denotes the mean absolute difference between designated property values and the property values of generated molecules. The result shows no significant difference of the three metrics between the two sets of scaffolds. This shows that our model achieves generalization over arbitrary scaffolds in generating valid molecules with controlled properties.

\begin{table}[htb!]
  \caption{Scaffold dependence of property-controlled generation. A scaffold is ``seen'' or ``unseen'' depending on it was in the training dataset or not.}
  \label{tbl:scaffold_dependence}
  \begin{tabular}{@{}l S[table-format=2.1] S[table-format=2.1] S[table-format=1.2]@{}}
    \toprule
    Property & {Validity} & {Uniqueness} & {MAD}  \\
    & {(\%)} & {(\%)} & \\
    \midrule
    MW (seen scaffolds) & 98.4 & 88.6 & 6.72  \\
    MW (unseen scaffolds) & 98.4 & 83.5 & 6.09  \\ \addlinespace
    TPSA (seen scaffolds) & 93.2 & 87.0 & 8.32  \\
    TPSA (unseen scaffolds) & 92.5 & 82.9 & 9.82  \\ \addlinespace
    LogP (seen scaffolds) & 98.2 & 91.1 & 0.28  \\
    LogP (unseen scaffolds) & 97.1 & 87.0 & 0.36  \\
    \bottomrule
  \end{tabular}
\end{table}

\subsection{Multi-property control}
\label{sec:multi-property control}
Designing new molecules seldom requires only one specific molecular property to be controlled.
Among others, drug design particularly involves simultaneous control of a multitude of molecular properties.
In this regard, we first tested our model's ability of simultaneously controlling two of MW, TPSA, and LogP.
We trained three instances of the model, each being jointly conditioned on MW and TPSA, MW and LogP, and LogP and TPSA.
We then specified each property with two target values (350 and 450 for MW, 50 and 100 for TPSA, and 2 and 5 for LogP) and combined them to prepare four generation conditions for each pair.
Under every generation condition, we used the randomly sampled 100 scaffolds that we used for the results of Secs.~\ref{sec:overall performance} and \ref{sec:single-property control} and generated 100 molecules from each scaffold.
We excluded those generations whose target MW is smaller than the used scaffold MW.

\begin{figure}[htb!]
    \centering
    \includegraphics[width=1.0\textwidth]{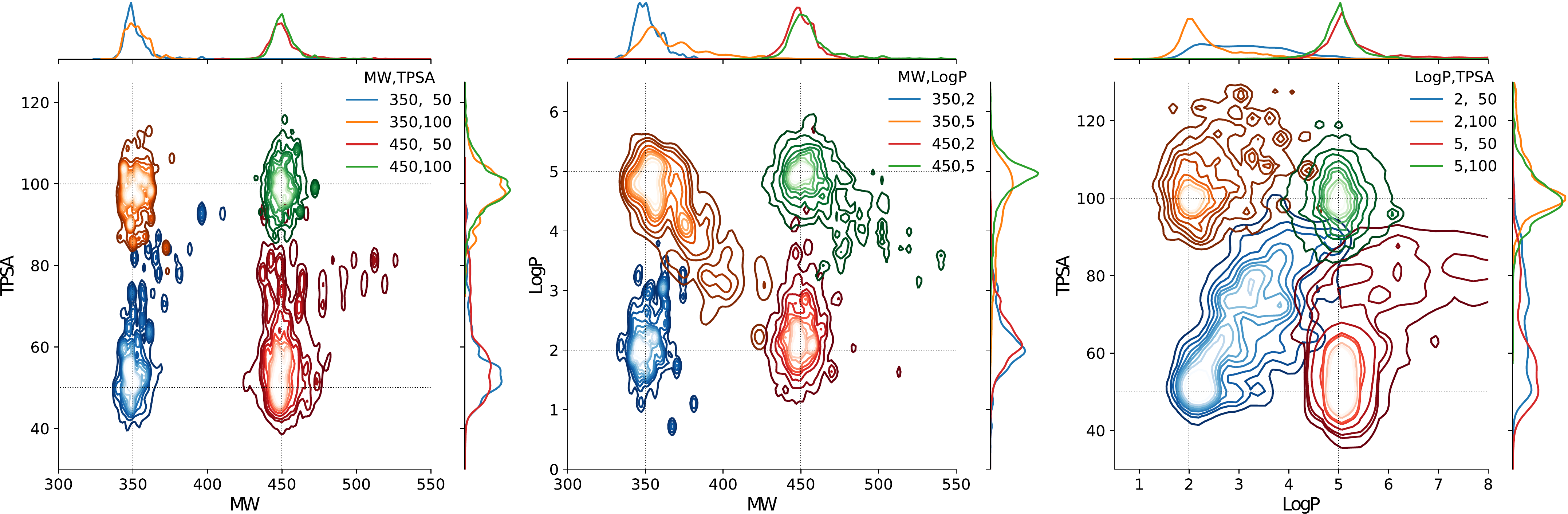}
    \caption{Estimated joint distributions of the property values of generated molecules. The legends show the target values used for the generations. In all distributions, the innermost contour encloses \SI{10}{\percent}, the outermost encloses \SI{90}{\percent}, and each $n$-th in the middle encloses $n\times$\SI{10}{\percent} of the population. On the upper and right ends of each plot are the marginal distributions of the abscissa and ordinate properties, respectively.}
    \label{fig:double_kdes}
\end{figure}

Figure~\ref{fig:double_kdes} shows the result of the generations conditioned on MW and TPSA, MW and LogP, and LogP and TPSA.
Plotted are the joint distributions of the property values over the generated molecules. 
Gaussian kernels were used for the kernel density estimation.
We see that the modes of the distributions are well located near the point of the target values.
As an exception, the distribution by the target (LogP, TPSA) = (2, 50) shows a relatively long tail over larger LogP and TPSA values.
This is because LogP and TPSA have by definition a negative correlation between each other and thus requiring a small value for both can make the generation task unphysical.
Intrinsic correlation in molecular properties can even cause seemingly feasible targets to result in dispersed property distributions. An example of such can be the result of another target (LogP, TPSA) = (5, 50), but we note that in the very case the outliers (in LogP > 5.5 and TPSA > 65 for example) amount to only a minor portion of the total generations, as the contours show.

\begin{figure}[hbt!]
    \centering
    \includegraphics[width=0.6\textwidth]{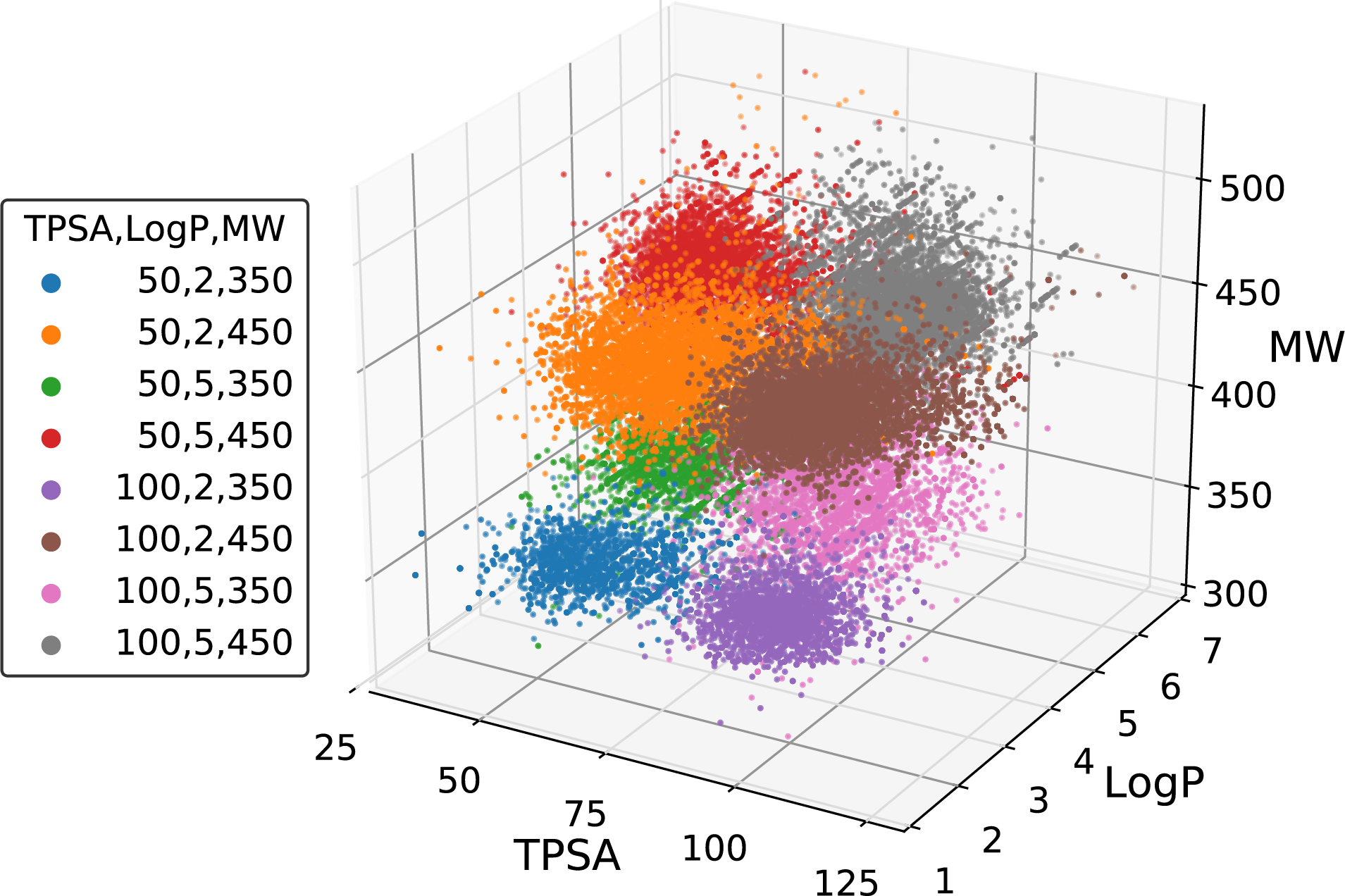}
    \caption{Scatter plot of the property values of generated molecules. The legend lists the eight sets of property values used for the generations.}
    \label{fig:triple_scatter}
\end{figure}

We further tested the conditioned generation by incorporating all the three properties.
We used the same target values of MW, TPSA, and LogP as above, resulting in total eight conditions of generation.
The rest of the settings, including the scaffold set and the number of generations, were retained.
The result is shown in Figure~\ref{fig:triple_scatter}, where we plotted the MW, TPSA, and LogP values of the generated molecules.
The plot shows that the distributions from different target conditions are well separated from one another.
As with the double-property result, all the distributions are well centered around their target values.

We also compared the generation performance of our model for single- and multi-property controls in a quantitative way.
Table~\ref{tbl:property_control_summary} shows the performance statistics of single-, double-, and triple-property controls in terms of MAD, validity, and novelty.
Using the same 100 scaffolds, we generated 100 molecules from each, each time under a randomly designated target condition.
As the number of incorporated properties increases from one to two and to three, the overall magnitudes of the descriptors are well preserved.
Regarding the slight increases in the MAD values, we attribute them to the additional confinement of chemical space forced by intrinsic correlations between multiple properties.
Nevertheless, the magnitudes of the worsening are small compared to the mean values of the properties (389 for MW, 77 for TPSA, and 3.6 for LogP).

\begin{table}[htb!]
  \caption{Statistical comparison of the performance on single-, double-, and triple-property controls.}
  \label{tbl:property_control_summary}
  \begin{tabular}{@{}l S[table-format=2.2] S[table-format=2.2] S[table-format=1.2] S[table-format=2.1] S[table-format=2.1]@{}}
    \toprule
    Properties & {MW MAD} & {TPSA MAD} & {LogP MAD} & {Validity} & {Novelty} \\
    & & & & {(\%)} & {(\%)} \\
    \midrule
    MW\textsuperscript{\emph{a}} & 6.53 & {--} & {--} & 98.3 & 98.7  \\
    TPSA\textsuperscript{\emph{a}} & {--} & 7.72 & {--} & 93.7 & 99.1 \\
    LogP\textsuperscript{\emph{a}} & {--} & {--} & 0.40 & 97.1 & 99.2 \\
    \addlinespace
    MW \& TPSA & 8.04 & 7.06 & {--} & 93.5 & 99.3 \\
    MW \& LogP & 11.59 & {--} & 0.45 & 97.0 & 99.5 \\
    TPSA \& LogP & {--} & 9.62 & 0.60 & 94.5 & 99.5 \\
    \addlinespace
    MW \& TPSA \& LogP & 16.23 & 10.95 & 0.73 & 93.9 & 99.7 \\
    \bottomrule
    \addlinespace
    \multicolumn{6}{@{}l@{}}{\textsuperscript{\emph{a}}From the result in Sec.~\ref{sec:overall performance}}
  \end{tabular}
\end{table}

\section{Conclusion}
\label{sec:Conclusion}

In this work, we proposed a scaffold-based molecular graph generative model.
The model generates new molecules that retain a desired substructure, i.e., scaffold, by sequentially adding new atoms and bonds to the graph of the scaffold.
In contrast to other related methods such as conditional generation, the strategy guarantees that the generated molecules naturally have the scaffold as a substructure.


We evaluated our model by examining the validity, uniqueness, and novelty of generated molecules. 
Despite the constraint on the search space imposed by scaffolds, the model showed comparable results with regard to previous SMILES-based and graph-based molecular generative models.
Our model consistently worked well in terms of the three metrics when new scaffolds which were not in the training set were given. 
This means that the model achieved good generalization rather than memorizing the pairings between the scaffolds and molecules in the training set. 
In addition, while retaining the given scaffolds, our model successfully generated new molecules with desirable degrees of molecular properties such as molecular weights, topological polar surface areas, and octanol-water partition coefficients.
The property-controlled generation could also incorporate multiple molecular properties simultaneously. We believe that our scaffold-based molecular graph generative model provides a practical way of optimizing the functionality of molecules with fixed core structures.


\section*{Appendix. Implementation details}
\label{sec:appendix}

We here describe our implementation of the model. The full process of encoding and decoding is shown in Algorithm~\ref{alg:ggm}.

\begin{algorithm}
    \caption{Scaffold-based graph generation}
    \label{alg:ggm}
    \hspace*{\algorithmicindent}\textbf{Inputs:} $G$, $S$, $\mathbf{y}$, $\mathbf{y}_S$ \Comment{Whole/scaffold graphs and properties}
\begin{algorithmic}[1]
    \State $G_0 \gets S$
    \State $\tilde{\mathbf{y}} \gets \mathrm{concat}\left( \mathbf{y}, \mathbf{y}_S \right)$
    \If{ $G\neq \left(\emptyset,\emptyset\right)$ } \Comment{Learning phase}
        \State  $\left(\mathbf{H}_{V(G)}, \mathbf{H}_{E(G)} \right)\gets \mathsf{embed}(G)$
        \State $\mathbf{H}_{V(G)} \gets \mathsf{propagate}^{(k)} \left(\mathbf{H}_{V(G)}, \mathbf{H}_{E(G)}, \tilde{\mathbf{y}} \right)$
        \State $\mathbf{z} \sim \mathsf{reparam}\circ\mathsf{readout}\left(\mathbf{H}_{V(G)}\right)$ \Comment{Vector representation of the target graph}
    \Else
        \State $\mathbf{z} \sim \mathcal{N}\left(\mathbf{0},\mathbf{I}\right)$ \Comment{Generation phase}
    \EndIf
    \State $\tilde{\mathbf{z}} \gets \mathrm{concat}\left(\mathbf{z}, \tilde{\mathbf{y}}\right)$
    \State $\left(\mathbf{H}_{V(G_0)}, \mathbf{H}_{E(G_0)}\right)\gets \mathsf{embed}\left(G_0\right)$ \Comment{Node and edge feature vectors}
    \State $\mathbf{H}_{V(G_0)} \gets \mathsf{propagate}^{(k)} \left(\mathbf{H}_{V(G_0)}, \mathbf{H}_{E(G_0)}, \tilde{\mathbf{y}} \right)$ \Comment{Initial update of the scaffold nodes}
    \State $t\gets 1$ \Comment{Node addition counter}
    \State $v_t \sim \mathrm{Cat}\circ\mathsf{addNode}\left(\mathbf{H}_{V(G_{t-1})}, \mathbf{H}_{E(G_{t-1})}, \tilde{\mathbf{z}} \right)$ \Comment{Sample an node type or $\mathsf{STOP}$}
    \While{ $v_t\neq\mathsf{STOP}$ }
        \State $V(G_t) \gets V(G_{t-1}) \cup \left\{v_t\right\}$ \Comment{Add the new node}
        \State $\mathbf{H}_{V(G_t)} \gets \mathbf{H}_{V(G_{t-1})} \cup \mathsf{initNode}\left(v_t, \mathbf{H}_{V(G_{t-1})}\right)$ \Comment{Initialize and add a new node vector}
        \State $E_{t,0} \gets E(G_{t-1})$; $\mathbf{H}_{E_{t,0}} \gets \mathbf{H}_{E(G_{t-1})}$ \Comment{Prepare edge additions}
        \State $i\gets1$ \Comment{Edge addition counter}
        \State $e_{t,i} \sim \mathrm{Cat}\circ\mathsf{addEdge}\left(\mathbf{H}_{V(G_t)}, \mathbf{H}_{E_{t,i-1}}, \tilde{\mathbf{z}} \right)$ \Comment{Sample an edge type or $\mathsf{STOP}$}
        \While{ $e_{t,i}\neq\mathsf{STOP}$ }
            \State $v_{t,i} \sim \mathrm{Cat}\circ\mathsf{selectNode} \left(\mathbf{H}_{V(G_t)}, \mathbf{H}_{E_{t,i-1}}, \tilde{\mathbf{z}} \right)$ \Comment{Sample a node to connect}
            \State $E_{t,i} \gets E_{t,i-1} \cup \left\{(v_t,v_{t,i})\right\}$ \Comment{Add the new edge (with type $e_{t,i}$)}
            \State $\mathbf{H}_{E_{t,i}} \gets \mathbf{H}_{E_{t,i-1}} \cup \mathsf{initEdge} \left(e_{t,i}, \mathbf{H}_{V(G_t)}\right)$ \Comment{Initialize and add a new edge vector}
            \State $i \gets i+1$
            \State $e_{t,i} \sim \mathrm{Cat}\circ \mathsf{addEdge} \left(\mathbf{H}_{V(G_t)}, \mathbf{H}_{E_{t,i-1}}, \tilde{\mathbf{z}} \right)$ \Comment{Sample a next edge type or $\mathsf{STOP}$}
        \EndWhile
        \State $\mathbf{H}_{E(G_t)}\gets \mathbf{H}_{E_{t,i-1}}$
        \State $E(G_t)\gets E_{t,i-1}$
        \State $G_t\gets \left(V(G_t), E(G_t)\right)$
        \State $t\gets t+1$
        \State $v_t\sim \mathrm{Cat}\circ \mathsf{addNode} \left(\mathbf{H}_{V(G_{t-1})}, \mathbf{H}_{E(G_{t-1})}, \tilde{\mathbf{z}} \right)$ \Comment{Sample a next node type or $\mathsf{STOP}$}
    \EndWhile
    \State $G_t^* \sim \mathrm{Cat}\circ \mathsf{selectIsomer}\left( G_t, \tilde{\mathbf{z}}\right)$ \Comment{Assign the stereoisomerism}
    \State \textbf{return} $G_t^*$
\end{algorithmic}
\end{algorithm}

\textbf{Graph representation of molecules.}
In our graph representation $G=\left(V(G),E(G)\right)$ of a molecule, the nodes $v \in V(G)$ represent the atoms, and the edges $(v,w) \in E(G)$ represent the bonds. We regard each node as attributed with an atom type and each edge as attributed with a bond type.

In the present work, we used the atom types in an indexed family $\mathcal{A} = \left(A_i\right) =$ (C, N, O, F, P, S, Cl, Br) and the bond types in another indexed family $\mathcal{B} = \left(B_i\right) =$ (single-bond, double-done, triple-bond). The symbols in $\mathcal{A}$ indicate the corresponding elements in the periodic table.
For the initial representation of whole-molecules and scaffolds, we use an extended family $\mathcal{A}^*$, which includes all the elements of $\mathcal{A}$ and additionally chirality (\textit{R}, \textit{S}, or none), formal charge, and aromaticity; also, we use an extended family $\mathcal{B}^*$, which includes the three bond types and stereoisomerism (\textit{E}, \textit{Z}, \textit{cis}, \textit{trans}, or none). We used RDKit\cite{landrum2006rdkit} to preprocess molecules into graphs.

To prepare node feature vectors $\mathbf{h}_v$ and edge feature vectors $\mathbf{h}_{vw}$, we embed node and edge types via two networks:
\begin{align}
    \label{eq:appx:node embedding}
    \mathbf{h}_v &= \mathrm{MLP}^n\left( \mathbf{h}^{0*}_v\right) \\
    \label{eq:appx:edge embedding}
    \mathbf{h}_{vw} &= \mathrm{MLP}^e\left( \mathbf{h}^{0*}_{vw}\right).
\end{align}
$\mathbf{h}^{0*}_v$ is a raw feature vector representing the type of $v$ based on $\mathcal{A}^*$, and similarly $\mathbf{h}^{0*}_{vw}$ is a raw feature vector of $(v,w)$ based on $\mathcal{B}^*$. For each of $\mathrm{MLP}^n$ and $\mathrm{MLP}^e$, we used a single linear layer with output dimension 128. The result of embedding all elements of a graph $G$ becomes
\begin{equation}
    \label{eq:embedding}
    \left(\mathbf{H}_{V(G)}, \mathbf{H}_{E(G)}\right) = \mathsf{embed}\left(G\right).
\end{equation}
We use the same module $\mathsf{embed}$ to embed all whole-molecules, scaffolds, and stereoisomers.

\textbf{Graph propagation and readout.}
The graph propagation module
\begin{equation}
    \label{eq:appx:propagate}
    \mathbf{H}'_{V(G)} = \mathsf{propagate}\left( \mathbf{H}_{V(G)}, \mathbf{H}_{E(G)}, \mathbf{c}\right)
\end{equation}
consists of the following processes:
\begin{gather}
    \label{eq:appx:directional message}
    \mathbf{m}_{u\rightarrow v} = \mathrm{ReLU}\circ \mathrm{MLP}^m\circ \mathrm{concat}\left( \mathbf{h}_u, \mathbf{h}_v, \mathbf{h}_{uv}, \mathbf{c}\right) \\
    \label{eq:appx:aggregated message}
    \mathbf{m}_v = \sum_{u: (u,v)\in E(G)} \mathbf{m}_{u\rightarrow v} \\
    \label{eq:appx:update}
    \mathbf{h}'_v = \mathrm{GRUCell}\left( \mathbf{m}_v, \mathbf{h}_v \right),
\end{gather}
where $\circ$ is the function composition, $\mathrm{ReLU}$ is the rectified linear unit\cite{nair2010relu}, $\mathbf{c}$ is a condition vector, and $\mathrm{GRUCell}$ is a gated recurrent unit cell\cite{Cho2014} (accepting $\mathbf{m}_v$ as the input and $\mathbf{h}_v$ as the hidden state). For $\mathrm{MLP}^m$ we used one linear layer with output dimension 128. We had $\mathrm{MLP}^m$ and $\mathrm{GRUCell}$ use a different set of parameters in different rounds of iterated propagation.

The readout module summarizes node features via the gated pooling:
\begin{align}
    \mathbf{h}_G &= \mathsf{readout}\left( \mathbf{H}_{V(G)}\right) \nonumber \\
    \label{eq:appx:readout}
    &= \frac{1}{\left\lvert V(G)\right\rvert} \sum_{v\in V(G)} \sigma\big( \mathrm{MLP}^r_2\left( \mathbf{h}_v\right) \big) \odot \mathrm{MLP}^r_1\left( \mathbf{h}_v\right),
\end{align}
where $\sigma$ is the sigmoid function, and $\odot$ is the elementwise product. For each of $\mathrm{MLP}^r_1$ and $\mathrm{MLP}^r_2$, we used a single linear layer.

We had $\mathsf{propagate}$ and $\mathsf{readout}$ in different modules have different sets of parameters. For instance, all $\mathsf{propagate}$ involved in $\mathsf{addNode}$, $\mathsf{addEdge}$, $\mathsf{selectNode}$, and $\mathsf{selectIsomer}$ have different $\mathrm{MLP}^m$ and $\mathrm{GRUCell}$. In addition, we used two different output dimensions for $\mathsf{readout}$: when reading-out the node features of any transient graph in a building process, we set the dimension equal to that of a node feature vector (i.e., 128); when encoding a whole-molecule graph, we set double.

\textbf{Encoding.}
With a whole-molecule graph $G$, we sample a latent vector $\mathbf{z}$ by applying the reparametrization trick:\cite{kingma2013auto}
\begin{align}
    \label{eq:appx:mu}
    \boldsymbol{\mu}_G &= \mathrm{MLP}^\mu\left( \mathbf{h}_G\right) \\
    \label{eq:appx:sigma}
    \boldsymbol{\sigma}_G &= \exp\left\{ \mathrm{MLP}^\sigma\left( \mathbf{h}_G\right) /2 \right\} \\
    \boldsymbol{\epsilon} &\sim \mathcal{N}\left(\mathbf{0}, \mathbf{I}\right) \\
    \mathbf{z} &= \boldsymbol{\mu}_G + \mathbf{\sigma}_G \odot \boldsymbol{\epsilon},
\end{align}
where $\mathcal{N}\left(\mathbf{0}, \mathbf{I}\right)$ is the standard normal distribution. In the last line we omitted the graph dependence of $\mathbf{z}$ for simplicity. For each of $\mathrm{MLP}^\mu$ and $\mathrm{MLP}^\sigma$, we used one linear layer with output dimension 128. Note that in Algorithm~\ref{alg:ggm} we used $\mathsf{reparam}$ to concisely express the sampling process.

\textbf{Decoder modules.}
The node addition module computes atom type probabilities as
\begin{align}
    \hat{\mathbf{p}}^{an} &= \mathsf{addNode}\left( \mathbf{H}_{V(G_t)}, \mathbf{H}_{E(G_t)}, \mathbf{z}\right) \nonumber \\
    \label{eq:appx:addNode}
    &= \mathrm{softmax}\circ \mathrm{MLP}^{an}\circ \mathrm{concat}\Big( \mathsf{readout}\circ \mathsf{propagate}^{(k)} \left(\mathbf{H}_{V(G_t)}, \mathbf{H}_{E(G_t)}, \mathbf{z}\right), \mathbf{z}\Big),
\end{align}
where $G_t$ is the transient graph in a building process. For $\mathrm{MLP}^{an}$, we used three linear layers with $\mathrm{ReLU}$ activations. The output dimensions of the layers were all 128. The length of the vector $\hat{\mathbf{p}}^{an}$ is $\left\lvert\mathcal{A}\right\rvert + 1$. The computed probabilities define a categorical distribution ($\mathrm{Cat}$), from which we sample an index $i$ such that
\begin{equation}
    i \sim \mathrm{Cat}\left( \hat{\mathbf{p}}^{an} \right) \quad (1\le i \le \left\lvert\mathcal{A}\right\vert + 1).
\end{equation}
If $i\le\left\lvert\mathcal{A}\right\vert$, the model adds a new node with the $i$-th chemical element $A_i$, or else the building process terminates.

If a new node $w$ is to be added to a transient graph $G_t$, the node initialization module computes a corresponding feature vector as follows:
\begin{align}
    \mathbf{h}_w &= \mathsf{initNode}\left( w, \mathbf{H}_{V(G_t)} \right) \nonumber \\
    \label{eq:appx:initNode}
    &= \mathrm{MLP}^i_2 \circ \mathrm{concat}\Big( \mathsf{readout}\left(\mathbf{H}_{V(G_t)}\right), \mathrm{MLP}^i_1 \left(\mathbf{h}^0_w\right) \Big).
\end{align}
In the last line, $\mathbf{h}^0_w$ is a raw feature representing the new node's type based on $\mathcal{A}$ (note the absence of an asterisk, unlike the one in Eq.~\ref{eq:appx:node embedding}). We used one linear layer for each of $\mathrm{MLP}^i_1$ and $\mathrm{MLP}^i_2$.

The edge addition module $\mathsf{addEdge}$ computes $\hat{\mathbf{p}}^{ae}$ in the same way as Eq.~\ref{eq:appx:addNode} but with an MLP of different weights. If the sampled index $i \sim \mathrm{Cat}\left( \hat{\mathbf{p}}^{ae} \right)$ 
is less than or equal to $\left\lvert\mathcal{B}\right\vert$, the model adds a new edge with the $i$-th bond type $B_i$. If $i = \left\lvert\mathcal{B}\right\rvert +1$, the model stops the edge addition.

To describe the node selection, let us suppose a new node $w$ was added to a transient graph $G_{t-1}$ so that $V(G_t) = V(G_{t-1})\cup w$ and $E(G_t) = E(G_{t-1})$. If a new edge is determined to be added, the module $\mathsf{selectNode}$ first updates the node features as
\begin{equation}
    \mathbf{H}'_{V(G_{t-1})} = \mathsf{propagate}^{(k)}\left( \mathbf{H}_{V(G_{t-1})}, \mathbf{H}_{E(G_{t-1})}, \mathbf{z}\right)
\end{equation}
and computes the selection probability for each existing node through the following steps:
\begin{gather}
    \hat{p}_u^{sn'} = \mathrm{MLP}^{sn}\circ \mathrm{concat}\left( \mathbf{h}'_u, \mathbf{h}_w, \mathbf{z}\right) \quad \forall u\in V(G_{t-1}) \\
    \hat{\mathbf{p}}^{sn} = \mathrm{softmax}\left( \hat{\mathbf{p}}^{sn'}\right).
\end{gather}
Then from the resulting categorical distribution $\mathrm{Cat}\left( \hat{\mathbf{p}}^{sn} \right)$, the model samples a node and connects it with $w$ (i.e., add the resulting edge to $E(G_t)$).

The edge initialization module $\mathsf{initEdge}$ computes edge feature vectors in the same way as Eq.~\ref{eq:appx:initNode}. The differences are that different MLPs are used and that $\mathbf{h}^0_{w}$ is replaced by a raw representation of the chosen bond type.

A complete specification of a molecular graph should assign the extended types in $\mathcal{A}^*$ and $\mathcal{B}^*$ to its elements. Motivated by the strategy of Jin et al.,\cite{Jin2018} our model assigns only the basic types in $\mathcal{A}$ and $\mathcal{B}$ during graph building and specifies stereoisomerism at the final stage of generation. Given a graph $G$, the isomer selection module $
\mathsf{selectIsomer}$ prepares the set of all possible stereoisomers of $G$ enumerated by RDKit. The graphs in the resulting set $\mathcal{I}(G)$ consist of nodes and edges that are fully typed according to $\mathcal{A}^*$ and $\mathcal{B}^*$. For each isomeric graph $I\in\mathcal{I}(G)$, $
\mathsf{selectIsomer}$ estimates the selection probability through
\begin{align}
    \left(\mathbf{H}_{V(I)}, \mathbf{H}_{E(I)}\right) &= \mathsf{embed}\left(I\right) \\
    \mathbf{H}'_{V(I)} &= \mathsf{propagate}^{(k)}\left( \mathbf{H}_{V(I)}, \mathbf{H}_{E(I)}, \mathbf{z}\right) \\
    \mathbf{h}_{I} &= \frac{1}{\left\lvert V(I)\right\rvert} \sum_{v\in V(I)} \mathbf{h}'_{v} \\
    \hat{p}_{I}^{si} &= \sigma\circ \mathrm{MLP}^s\circ \mathrm{concat}\left( \mathbf{h}_{I}, \mathbf{z}\right).
\end{align}

Note that multiple $I\in\mathcal{I}(G)$ can be valid for one $G$. For instance, there can be some $G$ whose stereocenters are only partially labelled by its data source, and in such case isomers with different labels on the same unlabelled stereocenters can all be regarded as valid. When generating a new molecule, however, we want our model to predict one isomer without ambiguity. Therefore, in the generation phase, the model normalizes the probabilities $\hat{p}_I^{si}$ by $\sum_I \hat{p}_I^{si}$ and then chooses one plausible isomer from the resulting categorical distribution.

\textbf{Learning.}
We used the molecule dataset described in Sec.~\ref{sec:datasets and experiments} for learning. The dataset consists of a set of scaffold molecules, $\mathcal{S}$, and a collection of each scaffold's whole-molecules, $\mathcal{D}(\mathcal{S}) = \left\{\mathcal{D}(S): S\in\mathcal{S}\right\}$, where $\mathcal{D}(S)$ is a set of whole-molecules of scaffold $S$. We can arrange $\mathcal{S}$ and $\mathcal{D}(\mathcal{S})$ into an indexed family $\left( (S_i, G_i) \right)_{1\le i\le \sum\left\lvert\mathcal{D}(S)\right\rvert}$, each of whose element is a pair of a scaffold and one of its whole-molecules. There can be duplicates of scaffolds or whole-molecules over different indices, but each pair $(S_i,G_i)$ is unique.

The individual loss $l_i$ due to each pair $(S_i,G_i)$ is a weighted sum of three losses: the graph building loss $l_i^\text{build}$, the isomer selection loss $l_i^\text{isomer}$, and the posterior approximation loss $l_i^\text{KL}$.
The second of the three is set to be
\begin{equation}
    \label{eq:appx:isomer loss}
    l_i^\text{isomer} = -\sum_{I\in\mathcal{I}\left(G_i\right)} \left( p^{si}_I \log\left( \hat{p}_I^{si}\right) + \left(1-p^{si}_I\right) \log\left(1-\hat{p}_I^{si}\right) \right),
\end{equation}
where $p^{si}_I$ is the true probability of selecting $I$.
The third of the three reads\cite{kingma2013auto}
\begin{equation}
    \label{eq:appx:kl loss}
    l_i^\text{KL} = -\frac{1}{2} \sum_j\left( 1 + \log\left(\sigma_{G_i,j}^2\right) - \mu_{G_i,j}^2 - \sigma_{G_i,j}^2 \right),
\end{equation}
where $\mu_{G_i,j}$ and $\sigma_{G_i,j}$ are the $j$-th elements of $\boldsymbol{\mu}_{G_i}$ and $\boldsymbol{\sigma}_{G_i}$, respectively (Eqs.~\ref{eq:appx:mu} and \ref{eq:appx:sigma}).

To describe the graph building loss, let us express the stepwise transitions from $S_i$ to $G_i$ by a finite sequence $\left(G_{i,0}, G_{i,1}, \cdots, G_{i,T}\right)$, where $G_{i,0}=S_i$ and $G_{i,T}=G_i$. The transition from $G_{i,t}$ to $G_{i,t+1}$ conforms to the true probability vector $\mathbf{p}_{i,t}$, which has one unity value for the correct building action and zeros for the others. During learning, the model reconstructs each $G_i$ from $S_i$ by estimating a sequence $\left(\hat{\mathbf{p}}_{i,0}, \hat{\mathbf{p}}_{i,1}, \cdots, \hat{\mathbf{p}}_{i,T-1}\right)$. Then the individual graph building loss can be defined by
\begin{equation}
    \label{eq:appx:build loss}
    l_i^\text{build} = -\sum_t \mathbf{p}_{i,t}\cdot \log\left( \hat{\mathbf{p}}_{i,t}\right).
\end{equation}

We minimized $\sum_i \left( l_i^\text{build}+ l_i^\text{isomer}+ \beta l_i^\text{KL}\right)$ to optimize our model and maximize the log-likelihood in Eq.~\ref{eq:objective}. We used 0.1 for the weight $\beta$. As for $k$, the number of iterations of $\mathsf{propagate}$, we set $k=3$ for the initial propagation of whole-molecule graphs and scaffold graphs and $k=2$ for $\mathsf{addNode}$, $\mathsf{addEdge}$, and $\mathsf{selectNode}$.

Finally, we remark the effect of node and edge orderings. Sequential generation of graphs requires their elements to be ordered. Different orderings amount to different sequences of graph transitions for the same $\left(S_i,G_i\right)$ pair. Similarly to Li et al.,\cite{Li2018} we trained two models using a fixed ordering for one and using random orderings for the other. We evaluated the two models in terms of the descriptors used in Sec.~\ref{sec:results and discussion} and confirmed that different orderings cause no significance change in performance. Therefore, we used a fixed ordering (assigned by RDKit when reading SMILES data) for all the results in Sec.~\ref{sec:results and discussion}.

\begin{acknowledgement}
This work was supported by the National Research Foundation of Korea (NRF) grant funded by the Korea government (MSIT)(NRF-2017R1E1A1A01078109).
We also thank Prof. Mu-Hyun Baik of the Center for Catalytic Hydrocarbon Functionalizations, Institute for Basic Science (IBS) and Department of Chemistry at KAIST, for providing the computing resources.

\end{acknowledgement}


\bibliography{ggm}

\end{document}